\documentclass[journal]{IEEEtran}
\usepackage[utf8]{inputenc}
\usepackage{url}
\usepackage{authblk}
\usepackage{graphicx}
\usepackage{color}
\usepackage{amsmath}
\usepackage{textcomp}
\usepackage{hyperref}
\usepackage{lipsum}
\usepackage{graphicx}
\usepackage{tikz}
\usepackage{caption}
\usepackage{subfigure}
\usepackage{cite}
\usepackage{adjustbox}
\usepackage{multirow}
\usetikzlibrary{positioning, shapes.geometric, arrows}

\tikzstyle{startstop} = [rectangle, rounded corners, minimum width=2cm, minimum height=1cm,text centered, draw=black, fill=white!30]

\tikzstyle{process} = [rectangle, minimum width=3cm, minimum height=1cm, text centered, draw=black, fill=orange!30]
\tikzstyle{decision} = [diamond, minimum width=3cm, minimum height=1cm, text centered, draw=black, fill=green!30]

\tikzstyle{arrow} = [thick,->,>=stealth]

\newcommand*{\bl}[1]{\textcolor{orange}{\textbf{BILAL}: #1}}
\newcommand*{\ws}[1]{\textcolor{blue}{\textbf{WS}: #1}}

\graphicspath{ {images/} }

\begin{document}

\title{Fake Visual Content Detection Using \\ Two-Stream Convolutional Neural Networks} 





\author{Bilal Yousaf, Muhammad Usama, Waqas Sultani, Arif Mahmood, Junaid Qadir
         
\thanks{B. Yousaf, W. Sultani, and  A. Mahmood are with Department
of Computer Science, Information Technology University, Lahore, Pakistan.  E-mails: \{msds18007, waqas.sultani, arif.mahmood\}@itu.edu.pk}
\thanks{M. Usama and J. Qadir are with Department of Electrical Engineering, Information Technology University, Lahore, Pakistan. E-mails: \{muhammad.usama, junaid.qadir\}@itu.edu.pk}}


\maketitle

\begin{abstract}
\sloppy
Rapid progress in adversarial learning has enabled the generation of realistic-looking fake visual content. To distinguish between fake and real visual content, several detection techniques have been proposed. The performance of most of these techniques however drops off significantly if the test and the training data are sampled from different distributions. This motivates efforts towards improving the generalization of fake detectors. Since current fake content generation techniques do not accurately model the frequency spectrum of the natural images, we observe that the frequency spectrum of the fake visual data contains discriminative characteristics that can be used to detect fake content. We also observe that the information captured in the frequency spectrum is different from that of the spatial domain. Using these insights, we propose to complement frequency and spatial domain features using a two-stream convolutional neural network architecture called TwoStreamNet. We demonstrate the improved generalization of the proposed two-stream network to several unseen generation architectures, datasets, and techniques. The proposed detector has demonstrated significant performance improvement compared to the current state-of-the-art fake content detectors and fusing the frequency and spatial domain streams has also improved generalization of the detector.

\end{abstract}


\section{Introduction}

\IEEEPARstart{R}{ecent} technological advancements in artificial intelligence (AI) have led to various beneficial applications in vision, language, and speech processing. However, at the same time, the power of these technologies may be exploited by adversaries for illegal or harmful uses. For example, Deepfakes---a portmanteau of the terms ``deep learning'' and ``fake''---may be used to produce or alter photo-realistic audio-visual content with the help of deep learning for an illegal or harmful purpose. Deepfake technology enables one to effectively synthesize realistic-looking fake audio or video of a real person speaking and performing in any arbitrary way \cite{chesney2019deep}. The term Deepfake was first coined by a Reddit community
for synthetically replacing the face of a person with the face of another person. The term expanded with time to include similar techniques such as Lip-Sync \cite{kumar2017obamanet, suwajanakorn2017synthesizing}, facial expression reenactment \cite{thies2016face2face, thies2015real, wiles2018x2face}, full-body and background manipulation as well as audio synthesis  \cite{chan2019everybody, cai2018deep, esser2018towards, wang2017tacotron, arik2018neural, tachibana2018efficiently}.


The rise of technology such as Deepfake has eroded the traditional confidence in the authenticity of audio and video as any digital content (audio, video, text) can be easily subverted using advanced deep learning techniques for synthesizing images trained on readily accessible public videos and images \cite{gogginbenjamin2018,leedave2018,celesamantha2018, deeptrace2018}. The gravity and urgency of the Deepfake threat can be gauged by noting that in recent times a CEO was scammed using Deepfake audio for \$243,000 \cite{damianiJ2019} and a fake video of the president of Gabon has resulted in a failed coup attempt. Other potential effects of the Deepfake threat include danger to journalism and democratic norms because elections can be manipulated and democratic discourse may be disrupted by creating fake speeches of contending leaders  \cite{chesney2019deep,suwajanakorn2017synthesizing}. Unfortunately, most of the current research focuses on creating and improving Deepfakes and there is a lack of focus on reliable  Deepfake detection. As reported in \cite{deeptrace2018}, 902 papers on Generative Adversarial Networks (GANs) were uploaded to the arXiv in 2018 but only 25 papers uploaded during the same time period related to the anti-forgery related topics.

Recent research shows that neural networks can be used for detecting fake content~\cite{bayar2016deep, cozzolino2017recasting, rahmouni2017distinguishing, rossler2018faceforensics, afchar2018mesonet}. These methods however require a large amount of fake and real training data to accurately learn the data distributions of both classes. The performance of these methods drops significantly on the unseen fake data sampled from a different distribution or generation process as the underlying network may overfit the training data and thereby lose its ability to generalize. The model can be further trained to classify previously unseen data but it will require a large amount of data from the new distribution which may not always be available in such problems. Attackers and defenders are continuously improving their approaches and rolling out new attacks and defenses. Therefore, it may be very difficult to collect a large amount of fake data for new manipulation techniques. For such scenarios, a fake content detector can detect fake content without even being explicitly trained on it. 

\begin{figure}[!t]
    \centering
    \includegraphics[width=9cm]{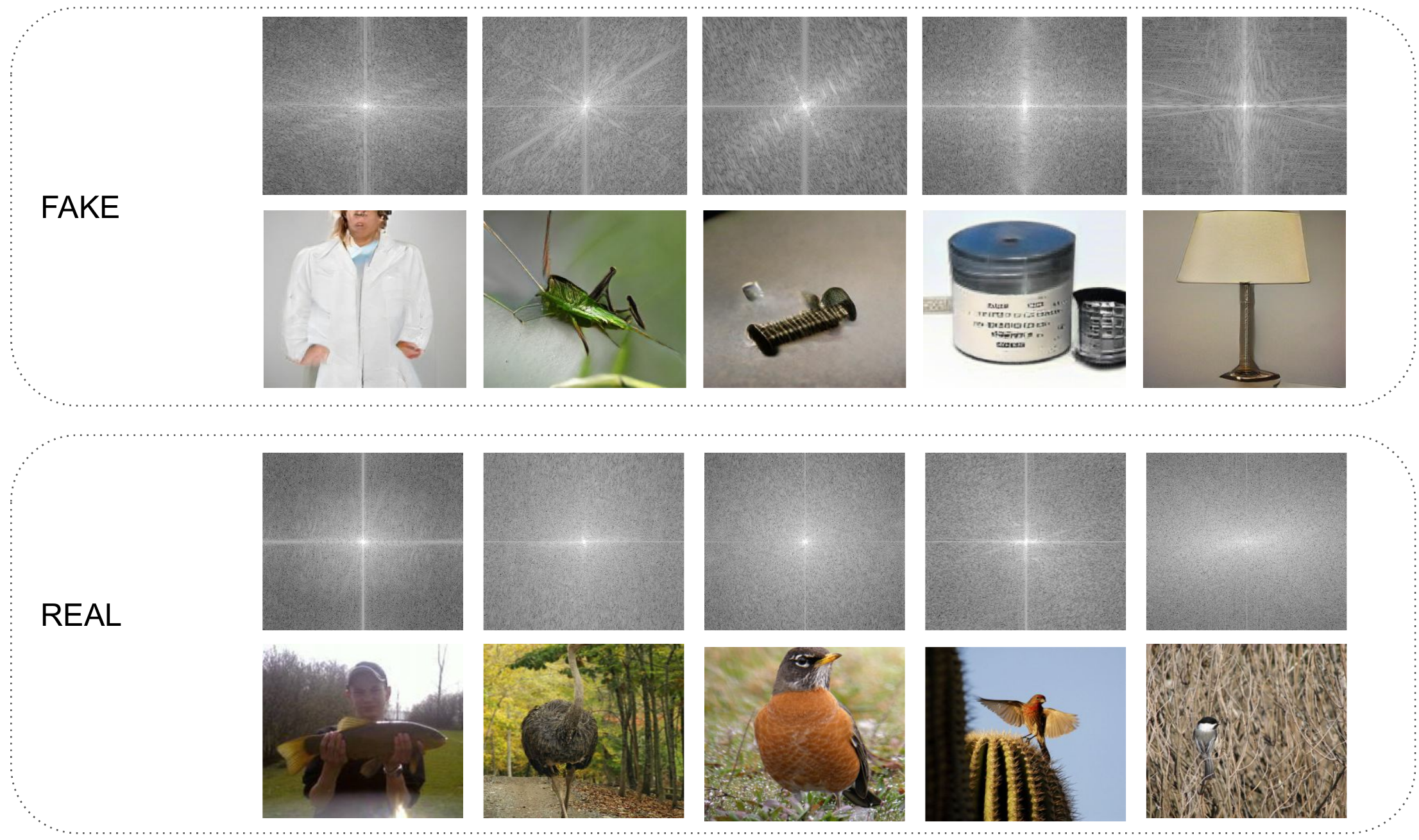}
    \caption{DFT-magnitude spectrum for fake and real images has discriminative features which can be exploited for improved fake detection performance.}
    \label{fig:spectrum}
\end{figure}

In the current work, we propose a two-stream network for fake visual content detection. The first stream called `Spatial Stream' detects the fake data employing RGB images while the second stream dubbed as `Frequency Stream' utilizes a combination of Discrete Fourier Transform (DFT) and Wavelet Transform (WT) for discriminating fake and real visual content.  
The frequency stream exploits the fact that the distribution of the frequency spectrum of the fake visual data remains distinct from the distribution of the real data frequency spectrum. This is illustrated in Figure \ref{fig:spectrum}, which shows the  DFT-magnitude spectrum for a sample of real and fake images. It can be seen that frequency spectrum has patterns that are different from that of real images. These differences are used to classify the fake versus real content. Since the information captured by the frequency stream is different from the information captured by the spatial stream, both these streams complement each other and fusing them together can provide better performance and generalization to unseen fake data detection. \textit{To the best of our knowledge, this is the first work that studies the fusion of cross-modal information fusion to improve fake content detection generalization}. 

The main contributions of this paper are summarized next.
\begin{enumerate}
    \item A novel two-stream architecture for fake visual content detection consisting of a Spatial Stream (SS) and a Frequency Stream (FS) is proposed. The SS learns the difference between the distributions of real and fake visual content in the spatial space using RGB images, while the FS learns to discriminate between the distributions of real and fake content in the frequency domain. The coefficients of the stationary frequencies are captured using DFT, while the coefficients of spatially varying multi-scale frequencies are captured using  Haar Wavelet transform. The spatial and frequency information complement each other therefore fusion of both has improved fake visual content detection.
    
    \item The proposed two-stream network comprising of a frequency and a spatial domain stream has outperformed the state-of-the-art fake detection methods with a significant margin. A detailed analysis of the proposed approach is performed and we empirically demonstrate that the proposed approach is robust across different quality JPEG compression and blurriness artifacts.
\end{enumerate}

In Section II, we discuss the related work and cover the traditional image forensics techniques and the latest deep learning-based image forensics algorithms with a prime focus on generalization. In Section III, we present our proposed methodology with all pre-processing schemes, training, and testing procedures. Section IV discusses the datasets used for evaluating and validating the proposed methodology. Section V provides the results and comprehensive evaluations of the generalization of the proposed methodology by performing an ablation study. Finally, Section VI concludes the paper and also points towards future directions.

\section{Related Work}
In this section, we briefly review recent  works needed to understand the state-of-the-art solutions in image forensics. We have divided this section into four subsections. We begin with a brief overview of the hand-crafted image forensic techniques followed by  deeply learned image forensic approaches. After that, we discuss methods that are focus to improve generalization. Finally, we conclude the section by covering the state-of-the-art frequency-domain techniques that are specifically designed for image forensic applications.


\subsection{Hand-Crafted Image Forensics}
A variety of methods are available in the literature for detecting traditional image manipulation techniques. Most of these manipulations are designed with the help of image editing tools. The traditional techniques make use of the hand-crafted features to detect specific clues that are created as a result of different manipulations. For example, several blind noise estimation algorithms have been proposed to detect region splicing forgeries \cite{agarwal2017photo,lyu2014exposing}. Popescu et al. \cite{popescu2005exposing} detected the image forgeries by estimating the resampling in the images. Haodong et al. \cite{7829291} integrated tampering possibility maps to improve forgery localization. Yuanfang et al. \cite{8293825} identified potential artifacts in hue, saturation, dark and bright channels of fake colorized images, and developed detection methods based on histograms and feature encoding. Similarly, Peng et al. \cite{8038041} used contact information of the standing objects and their supporting planes extracted from their reconstructed 3D poses to detect splicing forgeries. However, these techniques are unable to provide comparable performance to that of pixel-based methods in realistic situations. In recent works, learning-based techniques have become the preferred methods compared to traditional image forensics for achieving state-of-the-art detection performance \cite{zhou2016learning, huh2018fighting, cozzolino2015splicebuster, rao2016deep}.

\subsection{Deep Learning Based Image Forensics}
Due to the success of deep learning in different fields, several researchers have recently leveraged deep learning approaches for fake visual content detection. YanYang et al. \cite{8355817} proposed an algorithm based on difference images (DIs) and illuminant map (IM) as feature extractors to detect re-colorized images.  Quan et al. \cite{8355795} designed a deep CNN network with two cascaded convolutional layers to detect computer-generated images. McCloskey et al. \cite{mccloskey2018detecting} detected fake images by exploiting artifacts in the color cues, whereas Li et al. \cite{li2018exposing} used face warping artifacts for the forgery detection.  Li et al. \cite{li2018ictu} noticed that eye blinking in fake videos is different than the natural videos and used this fact to expose the fake videos. Similarly, Yang et al. \cite{yang2019exposing} have detected the Deepfakes by identifying the inconsistent head poses. Recently, Afchar et al.  \cite{afchar2018mesonet} proposed two compact forgery detection networks (Meso-4 and MesoInception-4) in which forgery detection is done by analyzing the mesoscopic properties\footnote{The eyes and mouth are determined as the mesoscopic features in the forgery detection in the Deepfake videos.} of Deepfake videos. Similarly, Nataraj et al. \cite{nataraj2019detecting} have shown that features extracted from the co-occurrence matrix can help improving fake data detection and Wang et al. \cite{wang2019fakespotter} proposed an anomaly detector based approach that uses pre-trained face detectors as a feature extractor. Although impressive, most of the above-mentioned approaches fail to perform well when fake visual data is sampled from a different distribution.

\subsection{Methods focused on Generalization}

In this subsection, we briefly describe the fake detection approaches focused on generalization. Cozzolino et al. \cite{cozzolino2018forensictransfer} proposed an auto-encoder based method to improve the performance of the model where learned weights are transferred for a different generation method. 
Zhang et al. \cite{zhang2019detecting} proposed a generalizable architecture named AutoGAN and evaluated its generalization ability on two types of generative networks. Xuan et al. \cite{xuan2019generalization} proposed that by using  Gaussian Blur or Gaussian noise, one can destroy unstable low-level noise cues and force models to learn more intrinsic features to improve the generalization ability of the model. Similarly, Wang et al. \cite{wang2019cnn} suggested that careful pre-and post-processing with data augmentation (such as blur and JPEG compression) improves the generalization ability. They have also shown improved fake detection results on multiple test sets by training on just one image generation network.

\subsection{Frequency Domain Methods}

Gueguen et al.~\cite{gueguen2018faster} extracted features from the frequency domain to perform classification tasks on images. Ehrlich et al.~\cite{ehrlich2019deep} proposed an algorithm to convert the convolutional neural network (CNN) models from the spatial domain to the frequency domain. Xu et al.~\cite{xu2020learning} proposed learning in the frequency domain and have shown that the performance of object detection and segmentation tasks get improved in the frequency domain as compared to using spatial RGB domain.
Durall et al.~\cite{durall2019unmasking} have shown that fake images have a difference in high-frequency coefficients compared to the natural images which he used  for fake detection. Wang et al.~\cite{wang2019cnn} have shown that the artifacts in the frequency spectrum of fake images  can be detected. Zhang et al.~\cite{zhang2019detecting} proposed that if instead of raw pixels, frequency spectrum (2D-DCT on all 3 channels) is used as an input to the fake image detector, the performance of the detector improves. These frequency response base detectors  target specific properties of the image generation process therefore, their performance degrades when  fake images from unseen distributions are tested. In contrast to these existing methods, the proposed algorithm fuses information from the spatial domain and the frequency domain to achieve improved generalization. Also, we propose to fuse DFT with Wavelet Transform to improve the discrimination in the frequency domain. These innovations have resulted in significant improvement in fake content detection compared to the existing methods.

\section{Methodology}

\begin{figure*}[!h]
    \centering
    \includegraphics[width=18cm]{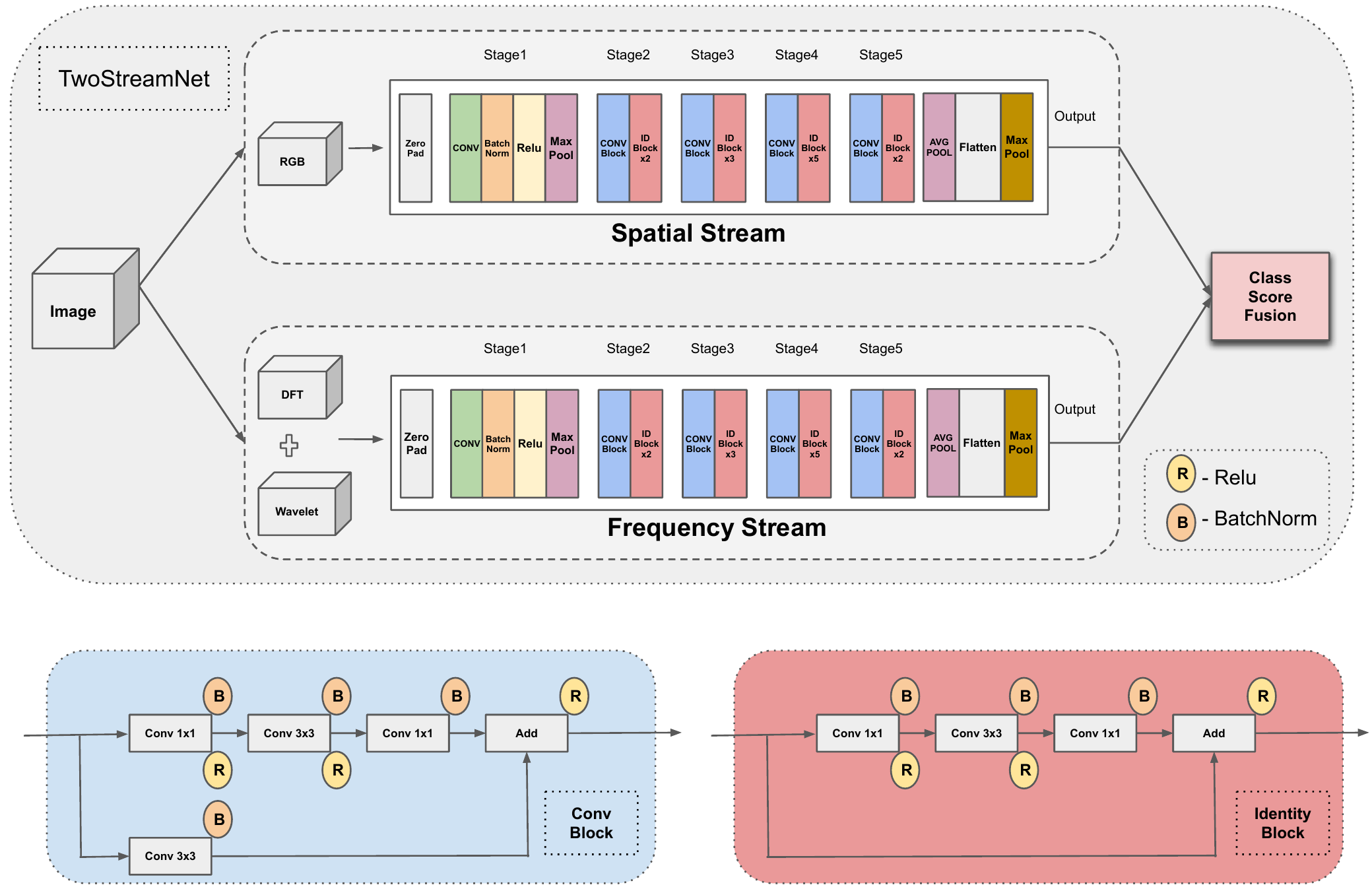}
    \caption{Proposed two-stream convolutional neural network (TwoStreamNet). The two network streams capture spatial and frequency domain artifacts separately, and their outputs are fused at the end of the network to produce classification scores.}
\label{fig: ensemble_full}
\end{figure*}

Improving the generalizability of a fake detection model is critical for its success in real-world applications where the fake content may be generated by unknown processes. We propose a generalizable fake detection model based on a two-stream convolutional network architecture shown in Fig. \ref{fig: ensemble_full}. 

The proposed architecture is motivated by the excellent performance of two-stream networks in action recognition in videos. To the best of our knowledge, The proposed network performs quite well on both seen and unseen data and has outperformed existing state-of-the-art (SOTA) methods in a wide range of experiments as we shall discuss in later sections. Our proposed two-stream network is novel and such a combination of frequency stream and the spatial stream has not been proposed before. In the following, we discuss the RGB to YCbCr conversion, DFT, DWT, and the proposed architecture in more detail.    


\subsubsection{The RGB to YCbCr Transformation}
The three channels in RGB color space are correlated with each other. We consider an orthogonal color space for improved representation. In our experiments, we have used YCbCr that has performed better than RGB space. 
As recommended in previous research \cite{bt2011studio} \cite{itu2002parameter}, the following formulas are used to convert from RGB to YCbCr color space:

\begin{equation}
\begin{split}
    Y = K_{ry} . R + K_{gy} . G + K_{by} . B,\\ 
    Cr = B - Y, \quad Cb = R - Y,\\
    \quad K_{ry} + K_{gy} + K_{by} = 1,
\end{split}
\end{equation}
where, $K_{ry}, K_{gy},$ and $K_{by}$ are the coefficients for color conversion whose values are specified in Table \ref{table:1} according to the standards. In our implementation, we used ITU601 \cite{ycbcr1}.
\begin{table}[h!]
\centering
\begin{tabular}{ |c|c|c| } \hline
\textbf{Reference Standard} & \textbf{$K_{ry}$} & \textbf{$K_{by}$} \\ \hline
\cite{ycbcr1} ITU601 / ITU-T 709 1250/50/2:1 & 0.299 & 0.114 \\ \hline
\cite{ycbcr2} ITU709 / ITU-T 709 1250/60/2:1 & 0.2126 & 0.0722 \\ \hline
\cite{smpte} SMPTE 240M (1999) & 0.212 & 0.087 \\ \hline
\end{tabular}
\caption{Coefficients $K_{ry}$ and $K_{by}$ of color conversion from RGB to YCbCr.}
\label{table:1}
\end{table}

\subsubsection{A Review of Frequency Domain Transforms}
\label{sec: dft_transform}
To fully capture the frequency information from a YCbCr image, we compute DFT and DWT for each image. 

\vspace{1mm}
\noindent\textit{Discrete Fourier Transform (DFT)}: Using DFT, one can decompose a signal into sinusoidal components of various frequencies ranging from 0 to maximum value possible based on the spatial resolution. For two dimensional data, i.e., images of size  $W \times H$, the DFT can be computed using the following formula:
\begin{multline*}
X_{w,h}=\sum_{n=0}^{W-1}\sum_{m=0}^{H-1} x_{w,h} e^{\frac{-i2\pi}{N}wn} e^{\frac{-i2\pi}{M}hm}, \hspace{10mm} (2) 
\end{multline*}
where $w$ is the horizontal spatial frequency, $h$ is the vertical spatial frequency, $x_{w,h}$ is the pixel value at coordinates (w, h), and $ X_{w, h}$ carries the magnitude and phase information of frequency at coordinates ($w, h$).

\vspace{1mm}
\noindent\textit{Discrete Wavelet Transform (DWT)}:  Wavelet transform decomposes an image into four different subband images.  High and low pass filters are applied at each row (column) and then they are downsampled by 2 to get the high and low-frequency components of each row (column) separately. In this way, the original image is converted into four sub-band images: High-high (HH), High-low (HL), Low-high (LH), and Low-low (LL). Each subband image preserves different features: HH region preserves high-frequency components in both horizontal and vertical direction, HL preserves high-frequency components in the horizontal direction and low-frequency components in the vertical direction, LH preserves low-frequency components in the vertical direction and high-frequency components in the horizontal direction and finally, LL preserves low-frequency components in the vertical direction and low-frequency components in the horizontal direction.


\subsection{Frequency Stream}
\label{sec: frequency_pre_processing}

\begin{figure*}[h!]
    \centering
    \includegraphics[width=17cm]{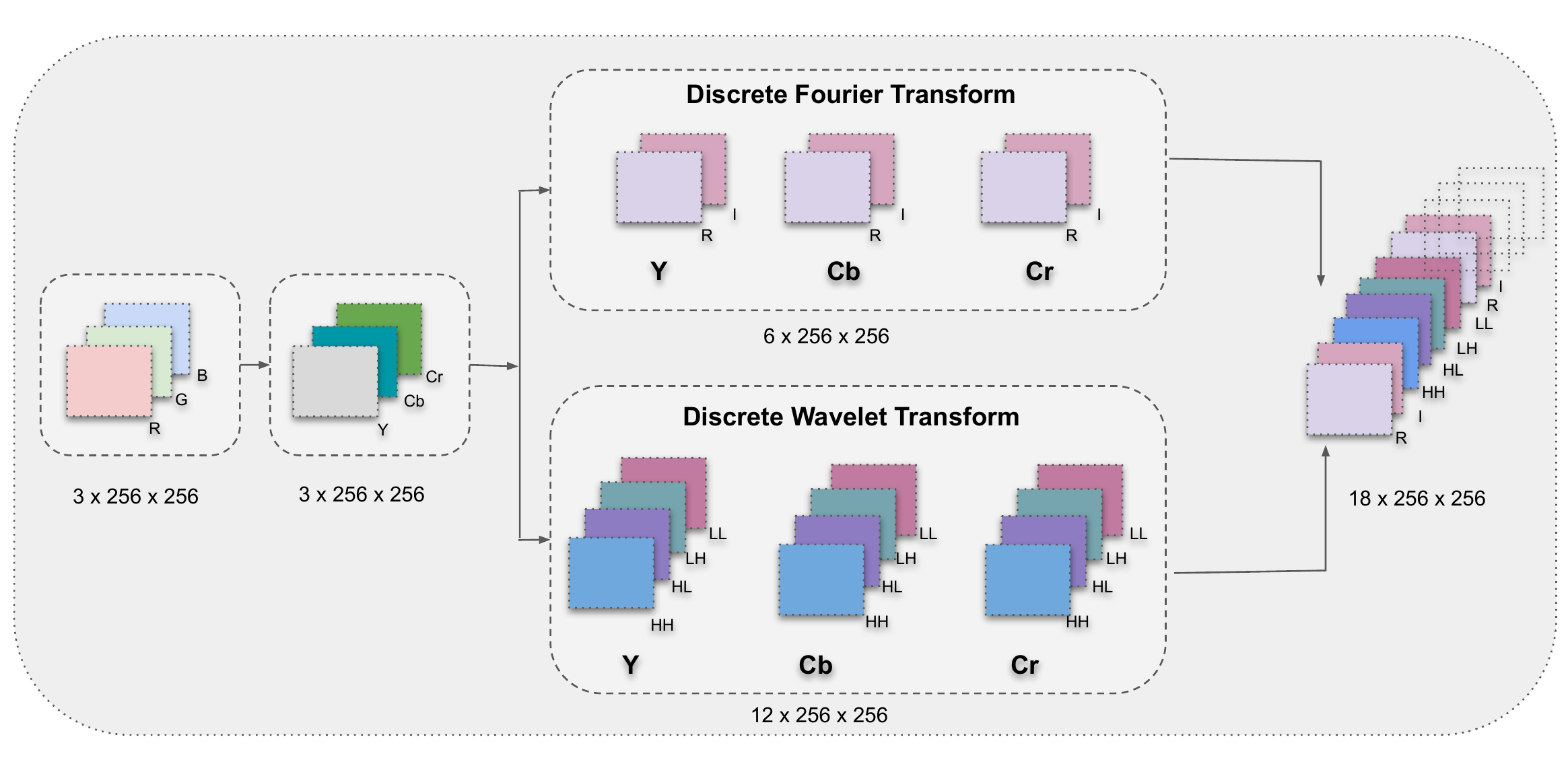}
    \caption{{Proposed pre-processing pipeline:} the input image is first converted to YCbCr color space and then transformed to the frequency domain by applying DFT and Wavelet Transforms (WT). After DFT, we get real (R) and imaginary (I) channels, and after WT we get four channels: HH, HL, LH, and LL.  The resulting channels are concatenated to form 3D cubes which are then input to the frequency stream for further processing.}
    \label{fig: FrequencyPreprocessing}
\end{figure*}

In this stream, two different types of the frequency spectrum are fused to get improved frequency domain representation which can better discriminate between the real and the fake visual content.  An overview of the frequency spectrum fusion is shown in Fig. \ref{fig: FrequencyPreprocessing}.  

The three YCbCr channels are then transformed to the frequency domain using two different types of transformations, including DFT and DWT. Each channel is divided into non-overlapping block of size 8 $\times$ 8 pixels and transformation is applied on each block independently. The resulting coefficients are then concatenated back to obtain the arrays of original image size.
The output of the DFT converts one input channel into two output channels corresponding to real and imaginary coefficients. Similarly, the DWT converts one input channel into 4 output channels corresponding to low frequencies (LL), high and low frequencies (HL), high frequencies (HH), and low and high frequencies (LH).  For three input channels (YCbCr), we obtain 18 output channels, 6 from DFT and 12 from DWT. All of these frequency output channels are concatenated to form 3D cubes of size $H \times W \times C$, where $H$ is the height and  $W$ is the width of the image, and $C$=18 are the number of channels.  We empirically observe that both DFT and DWT are necessary to capture essential information in the frequency domain at varying scales for improving the generalization ability of the proposed network.

\subsection{Spatial Stream}
\label{sec: spatialstream}

In this stream, RGB channels of the image are passed as input to the ResNet50 \cite{he2016deep} as the classifier. RGB images are augmented in a special way using JPEG compression and Gaussian Blur as recommended by Wang et al.~\cite{wang2019cnn}. This stream is trained individually and plugged in the TwoStreamNet at the test time. 



\subsection{Two Stream Network Architecture}
\label{ensemble}

The proposed two-stream network architecture is shown in Figure \ref{fig: ensemble_full}. ResNet50 network is used as a backbone in both of the streams of the proposed architecture. Since the number of input channels in the frequency stream is larger as compared to the spatial stream, therefore first layer of FS is accordingly modified. Both streams are independently trained and the output of both streams is fused using the class probability averaging fusion method.
In this fusion scheme, both streams contribute equally to the output, to produce the final classification probability. The performance of the combined scores is significantly better than the performance of the individual streams.  



\begin{table*}
\centering

\begin{tabular}{|c|c|c|c|c|c|c|c|c|c|c|c|c|c|}
\hline
\textbf{Metrics} & \textbf{Method} & \begin{tabular}[c]{@{}c@{}}\textbf{Star}\\\textbf{GAN} \end{tabular} & \begin{tabular}[c]{@{}c@{}}\textbf{Style}\\\textbf{GAN} \end{tabular} &
\textbf{SITD} & \begin{tabular}[c]{@{}c@{}}\textbf{Big}\\\textbf{GAN} \end{tabular} & \begin{tabular}[c]{@{}c@{}}\textbf{Style}\\\textbf{GAN2} \end{tabular} & \begin{tabular}[c]{@{}c@{}}\textbf{Cycle}\\\textbf{GAN} \end{tabular} & \begin{tabular}[c]{@{}c@{}}\textbf{Which}\\\textbf{face}\\\textbf{is}\\\textbf{real} \end{tabular} & 
\textbf{SAN} &
\begin{tabular}[c]{@{}c@{}}\textbf{Deep}\\\textbf{fake} \end{tabular} &
\begin{tabular}[c]{@{}c@{}}\textbf{Gua}\\\textbf{GAN} \end{tabular} &
\textbf{CRN} &
\textbf{IMLE}
\\ \hline \hline
\multirow{3}{*}{Accuracy} & \multicolumn{1}{|c|}{Wang et al. \cite{wang2019cnn}} & \multicolumn{1}{|c|}{$91.7$} & \multicolumn{1}{|c|}{$87.1$} & \multicolumn{1}{|c|}{$90.3$} & \multicolumn{1}{|c|}{$70.2$} & \multicolumn{1}{|c|}{$84.4$} & \multicolumn{1}{|c|}{$\textbf{85.2}$} & \multicolumn{1}{|c|}{$83.6$} & \multicolumn{1}{|c|}{$\textbf{53.5}$} & \multicolumn{1}{|c|}{$50.5$} & \multicolumn{1}{|c|}{$78.9$} & \multicolumn{1}{|c|}{$\textbf{86.3}$} & \multicolumn{1}{|c|}{$\textbf{86.2}$} \\ \cline{2-14} 

& \multicolumn{1}{|c|}{Frequency Stream (Ours)} & \multicolumn{1}{|c|}{$\textbf{97.67}$} & \multicolumn{1}{|c|}{$\textbf{89.36}$} & \multicolumn{1}{|c|}{$81.11$} & \multicolumn{1}{|c|}{$\textbf{72.08}$} & \multicolumn{1}{|c|}{$\textbf{91.83}$} & \multicolumn{1}{|c|}{$77.53$} & \multicolumn{1}{|c|}{$80.15$} & \multicolumn{1}{|c|}{$49.32$} & \multicolumn{1}{|c|}{$\textbf{68.90}$} & \multicolumn{1}{|c|}{$70.91$} & \multicolumn{1}{|c|}{$55.06$} & \multicolumn{1}{|c|}{$55.06$} \\ \cline{2-14}

& \multicolumn{1}{|c|}{Two Stream (Ours)} & \multicolumn{1}{|c|}{$\textbf{96.32}$} & \multicolumn{1}{|c|}{$\textbf{88.90}$} & \multicolumn{1}{|c|}{$\textbf{97.22}$} & \multicolumn{1}{|c|}{$\textbf{72.85}$} & \multicolumn{1}{|c|}{$\textbf{87.43}$} & \multicolumn{1}{|c|}{$84.09$} & \multicolumn{1}{|c|}{$\textbf{87.50}$} & \multicolumn{1}{|c|}{$50.23$} & \multicolumn{1}{|c|}{$\textbf{55.00}$} & \multicolumn{1}{|c|}{$\textbf{79.64}$} & \multicolumn{1}{|c|}{$77.75$} & \multicolumn{1}{|c|}{$77.75$} \\ \hline \hline

\multirow{3}{*}{\begin{tabular}[c]{@{}c@{}}F1-Score\\(Fake) \end{tabular}} & \multicolumn{1}{|c|}{Wang et al. \cite{wang2019cnn}} & \multicolumn{1}{|c|}{$91.31$} & \multicolumn{1}{|c|}{$85.19$} & \multicolumn{1}{|c|}{$89.91$} & \multicolumn{1}{|c|}{$61.13$} & \multicolumn{1}{|c|}{$81.53$} & \multicolumn{1}{|c|}{${84.2}$} & \multicolumn{1}{|c|}{$81.92$} & \multicolumn{1}{|c|}{$3.56$} & \multicolumn{1}{|c|}{$12.83$} & \multicolumn{1}{|c|}{$75.45$} & \multicolumn{1}{|c|}{$\textbf{87.93}$} & \multicolumn{1}{|c|}{$\textbf{87.88}$} \\ \cline{2-14} 

& \multicolumn{1}{|c|}{Frequency Stream (Ours)} & \multicolumn{1}{|c|}{$\textbf{97.63}$} & \multicolumn{1}{|c|}{$\textbf{88.21}$} & \multicolumn{1}{|c|}{$83.57$} & \multicolumn{1}{|c|}{$\textbf{71.71}$} & \multicolumn{1}{|c|}{$\textbf{91.14}$} & \multicolumn{1}{|c|}{$78.75$} & \multicolumn{1}{|c|}{$82.08$} & \multicolumn{1}{|c|}{$\textbf{28.39}$} & \multicolumn{1}{|c|}{$\textbf{61.13}$} & \multicolumn{1}{|c|}{$72.65$} & \multicolumn{1}{|c|}{$68.99$} & \multicolumn{1}{|c|}{$68.99$} \\ \cline{2-14}

& \multicolumn{1}{|c|}{Two Stream (Ours)} & \multicolumn{1}{|c|}{$\textbf{96.21}$} & \multicolumn{1}{|c|}{$\textbf{87.52}$} & \multicolumn{1}{|c|}{$\textbf{97.27}$} & \multicolumn{1}{|c|}{$\textbf{66.61}$} & \multicolumn{1}{|c|}{$\textbf{85.63}$} & \multicolumn{1}{|c|}{$\textbf{84.98}$} & \multicolumn{1}{|c|}{$\textbf{87.15}$} & \multicolumn{1}{|c|}{$4.39$} & \multicolumn{1}{|c|}{$\textbf{17.95}$} & \multicolumn{1}{|c|}{$\textbf{77.34}$} & \multicolumn{1}{|c|}{$81.79$} & \multicolumn{1}{|c|}{$81.79$} \\ \hline \hline

\multirow{3}{*}{\begin{tabular}[c]{@{}c@{}}F1-Score\\(Real) \end{tabular}} & \multicolumn{1}{|c|}{Wang et al. \cite{wang2019cnn}} & \multicolumn{1}{|c|}{$92.14$} & \multicolumn{1}{|c|}{$88.56$} & \multicolumn{1}{|c|}{$90.62$} & \multicolumn{1}{|c|}{$75.81$} & \multicolumn{1}{|c|}{$86.50$} & \multicolumn{1}{|c|}{${86.08}$} & \multicolumn{1}{|c|}{$85.0$} & \multicolumn{1}{|c|}{$\textbf{66.67}$} & \multicolumn{1}{|c|}{$68.28$} & \multicolumn{1}{|c|}{$81.5$} & \multicolumn{1}{|c|}{$\textbf{84.13}$} & \multicolumn{1}{|c|}{$\textbf{84.08}$} \\ \cline{2-14} 

& \multicolumn{1}{|c|}{Frequency Stream (Ours)} & \multicolumn{1}{|c|}{$\textbf{97.71}$} & \multicolumn{1}{|c|}{$\textbf{90.32}$} & \multicolumn{1}{|c|}{$77.78$} & \multicolumn{1}{|c|}{$72.43$} & \multicolumn{1}{|c|}{$\textbf{92.42}$} & \multicolumn{1}{|c|}{$76.97$} & \multicolumn{1}{|c|}{$77.76$} & \multicolumn{1}{|c|}{$60.78$} & \multicolumn{1}{|c|}{$\textbf{74.08}$} & \multicolumn{1}{|c|}{$68.93$} & \multicolumn{1}{|c|}{$18.41$} & \multicolumn{1}{|c|}{$18.41$} \\ \cline{2-14}

& \multicolumn{1}{|c|}{Two Stream (Ours)} & \multicolumn{1}{|c|}{$\textbf{96.43}$} & \multicolumn{1}{|c|}{$\textbf{90.01}$} & \multicolumn{1}{|c|}{$\textbf{97.18}$} & \multicolumn{1}{|c|}{$\textbf{77.13}$} & \multicolumn{1}{|c|}{$\textbf{88.84}$} & \multicolumn{1}{|c|}{$\textbf{86.20}$} & \multicolumn{1}{|c|}{$\textbf{87.83}$} & \multicolumn{1}{|c|}{$66.36$} & \multicolumn{1}{|c|}{$\textbf{69.00}$} & \multicolumn{1}{|c|}{$\textbf{81.5}$} & \multicolumn{1}{|c|}{$71.39$} & \multicolumn{1}{|c|}{$71.39$} \\ \hline

\end{tabular}

\caption{Comparison of the proposed Frequency Stream (FS) and Two-Stream network with the state-of-the-art method \cite{wang2019cnn} using average accuracy. Best results of Wang et al. \cite{wang2019cnn} with data augmentation using blur and JPEG (0.1) are reported where 0.1 mean JPEG compression is applied on 10\% images. The same augmentation is also used in the proposed approaches. Both our approach and that of Wang et al. are trained using ProGAN only and tested on the data generated by 12 unseen generation processes mentioned in the top row.} 

\label{table:cvpr_freq_twostream}
\end{table*}

\section{Experiments and Results}
\label{dataset}

\vspace{1mm}
\noindent\textbf{Training Dataset:} Following the protocol used by \cite{wang2019cnn}, the proposed two-stream network is trained using the fake images generated by  ProGAN \cite{karras2017progressive} and tested on the images generated by many other GANs.  ProGAN has 20 different officially released models trained on different object categories 
of the LSUN dataset, which is a large scale image dataset containing around one million labeled images for each of the 10 scene categories and 20 object categories\footnote{\url{https://www.yf.io/p/lsun}} \cite{yu2015lsun}. We choose 15 (airplane, bird, boat, bottle, bus, car, cat, chair, dog, horse, motorbike, person, sofa, train, and tv monitor) out of 20 models to create our validation and training set. We generated 10k fake images for training and 500 fake images for validation using each of the 15 models.  For each of these 15 categories of fake images, we collect 10k of real images for training and 500 for validation randomly from the LSUN dataset \cite{yu2015lsun}. In total, we have 300K training images and 15K validation images. For real images, we center crop the images equal to the size of the shorter edge and then resize the images to 256 $\times$ 256.

\vspace{1mm}
\noindent\textbf{Testing Dataset:} Testing dataset contains images which were generated using completely unseen generators as described in Table \ref{table:dataset_details}. To remain consistent with the current state of the art, the same generators are selected as that of \cite{wang2019cnn}.      The real images for testing purposes are obtained from the repository for each generator.
    



\begin{table}[h!]
\centering
\begin{tabular}{ |c|c|c| }
\hline
\textbf{Dataset} & \textbf{No. of Real Images} & \textbf{No. of Fake Images} \\ \hline
StarGAN \cite{choi2018stargan} & 1999  & 1999 \\ \hline
StyleGAN \cite{karras2019style} & 5991 & 5991 \\ \hline
SITD \cite{chen2018learning} & 180 & 180 \\ \hline
BigGAN \cite{brock2018large} & 2000 & 2000 \\ \hline
StyleGAN2 \cite{karras2019analyzing} & 7988  & 7988 \\ \hline
CycleGAN \cite{zhu2017unpaired} & 1321 & 1321 \\ \hline
Whichfaceisreal \cite{whichfaceisreal} & 1000 & 1000 \\ \hline
GauGAN  \cite{Park_2019_CVPR} & 5000 & 5000 \\ \hline
Deepfake  \cite{rossler2019faceforensics++} & 2698 & 2707 \\ \hline
CRN \cite{chen2017photographic} & 6382 & 6382 \\ \hline
IMLE \cite{li2019diverse} & 6382 & 6382 \\ \hline
SAN \cite{dai2019second} & 219 & 219 \\ \hline
\end{tabular}
\caption{Details of the testing dataset}.
\label{table:dataset_details}
\end{table}



\subsection{Implementation details}
\label{Implementation_details}

For training the FS, we use the Adam \cite{kingma2014adam} optimizer with an initial learning rate of 0.0001, weight decay of 0.0005, and a batch size of 24. 
 For all the training sets, we train the proposed network for 24 epochs.  Large training data has helped the model to converge quickly. Lastly, we select the best model based on the validation set. While training each stream, data augmentation based on Gaussian blur and JPEG compression with 10\% probability is used.

\subsection{Comparison with the Existing State-of-the-Art Algorithms}
\label{sec:Comparison}






\begin{figure*}[h!]
    \centering
    \includegraphics[width=17cm]{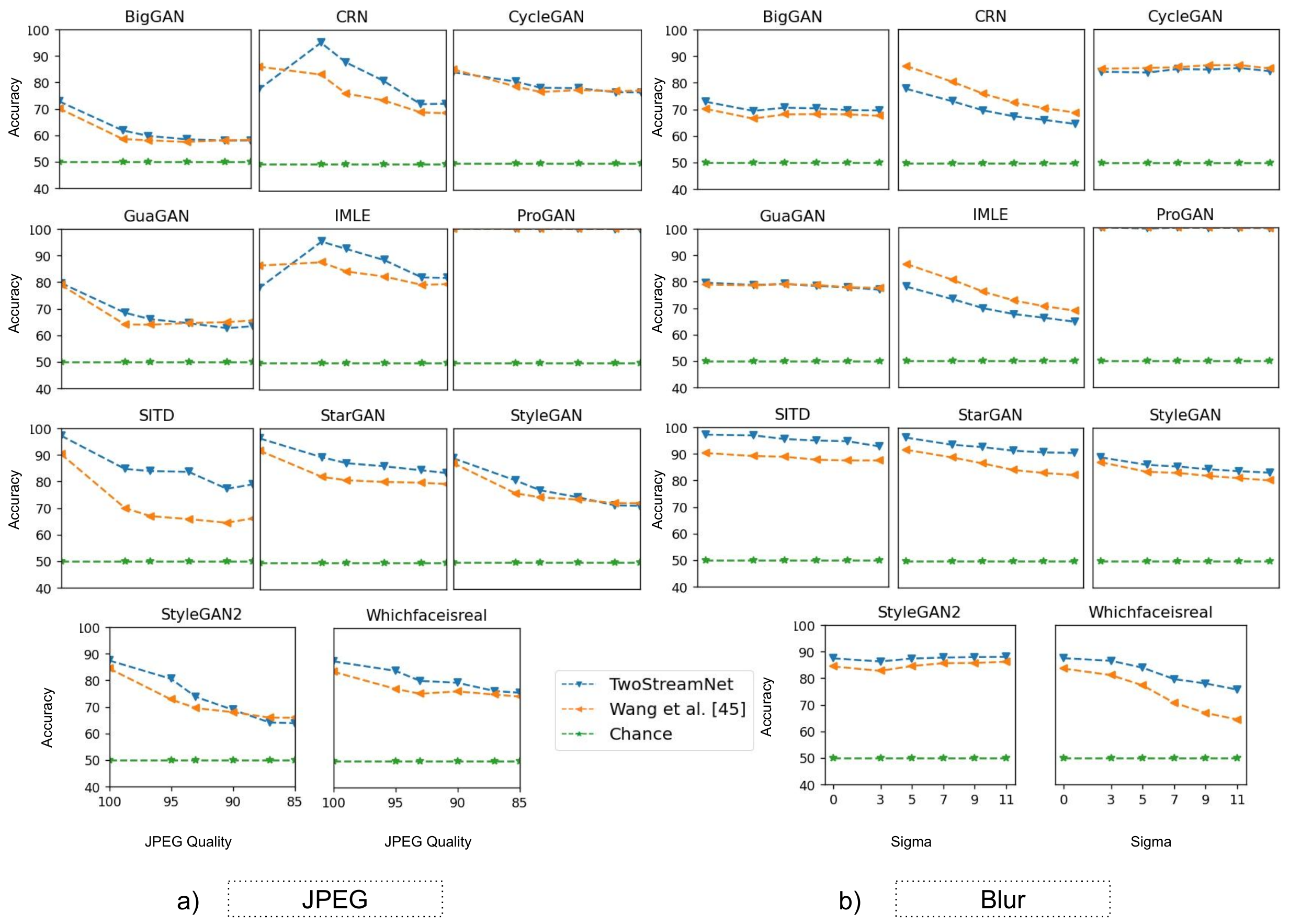}
    \caption{Robustness comparison of the proposed algorithm with Wang et al. \cite{wang2019cnn} for Gaussian Blur and JPEG Compression artifacts. In most experiments, the proposed two-stream net. We apply Gaussian Blur and JPEG Compression of different sizes on the test sets and measure the effect on the accuracy of our model. Our model performs near to the best for all the cross-modal datasets even when a large blurring effect is applied. Results show that our proposed solution is more robust as compared to the state of the art.}
    \label{fig:Robustness_Analysis}
\end{figure*}
\begin{figure*}[h!]
    \centering
    \includegraphics[width=17cm]{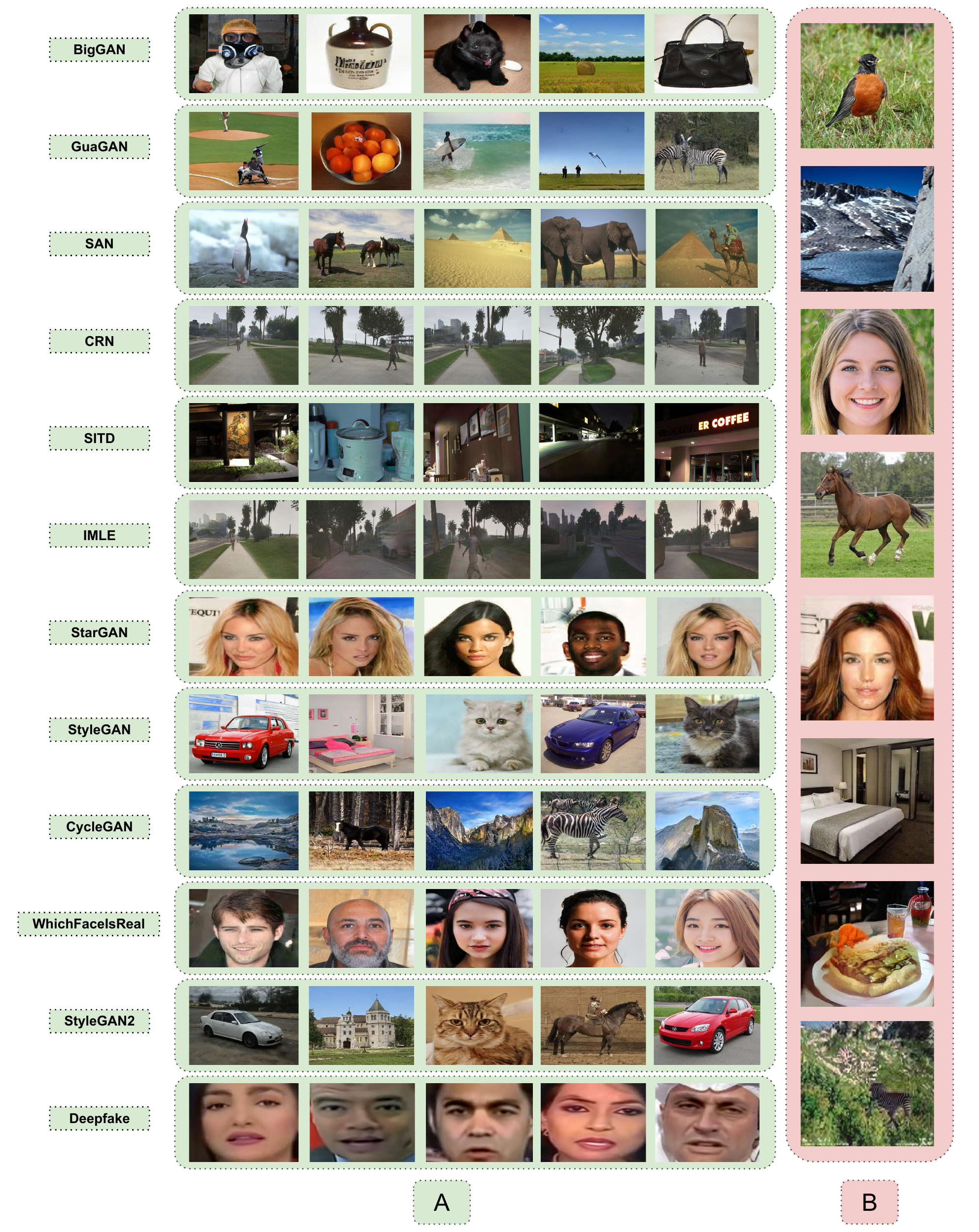}
    \caption{(A) Examples of fake images correctly detected by the proposed two-stream network, however, misclassified by Wang et al. \cite{wang2019cnn}; (B) Examples of fake image misclassified by both our proposed method and that of Wang et al. \cite{wang2019cnn}.} 
    \label{fig:final_detected}
\end{figure*}

We thoroughly evaluated the performance of the proposed method on the test dataset and compared it with the existing state of the art \cite{wang2019cnn}. We also compared the robustness analysis of our approach against some common real-world perturbations. In Table \ref{table:cvpr_freq_twostream}, we have shown a comparison of our results with the best results of Wang et al. \cite{wang2019cnn} ([Blur+JPEG(0.1)]). Their results from their official web link\footnote{\url{https://peterwang512.github.io/CNNDetection/}}. Results show that our  FS approach performs very well on the unseen manipulations and outperformed the state-of-the-art on several test sets while having competitive performance on the remaining. Results of the two-stream architecture demonstrate that our complete approach outperformed the state-of-the-art in almost all of the cases. Analysis of the results shows that when both spatial and frequency streams are combined into a two-stream architecture, they compliment each other in a way that their combined accuracy is greater than any of them individually. This clearly shows that FS ConvNet has learned distinctive features that were not learned by SS ConvNet. Overall, the combination of FS and SS plays a vital role in improving the generalization ability of the fake image detectors. In Figure \ref{fig:final_detected}(A), we have shown samples of the fake images which are misclassified by the state-of-the-art and are correctly classified by our proposed two-stream approach. These results demonstrate the ability of the proposed approach to detect high-quality fake images which are even very hard to discriminate by humans. Figure \ref{fig:final_detected}(B) shows the fake images which are misclassified by both Wang et al. \cite{wang2019cnn} and us.

\vspace{1mm}
\noindent\textbf{Robustness Analysis:}
In real-world settings, fake images may undergo several post-processing operations like compression, smoothness, etc. Therefore, we have evaluated the performance of the proposed model on the images which are post processed using JPEG compression and Gaussian blur. Specifically, we apply Gaussian blur with different standard deviations including [3, 5, 7, 9, 11] and JPEG compression with JPEG image quality factor of [85, 87, 90, 92, 95]. Results in Figure \ref{fig:Robustness_Analysis} show that our approach is robust to common perturbations. For most of the models, the proposed approach significantly outperformed the state-of-the-art method at varying blur levels. Similarly, the proposed approach also performed better than the state-of-the-art methods for a wide range of JPEG compressions.

\section{Ablation Study}
\label{sec:exp}

In this section, we thoroughly validate the different components of the proposed approach by performing an ablation study.

\subsection{Combining DFT and DWT}
As shown in Figure \ref{fig: ensemble_full}, we propose to combine DWT and DFT for better feature representation and robust fake content detection. To verify the effectiveness of using both transformations, while keeping all the experimental settings the same, we experimented with DFT and DWT separately.  After training for 20 epochs the best epoch is chosen based on validation data accuracy. The results shown in  Table \ref{table:dft_dwt_results} demonstrate that a combination of DFT and DWT is essential to produce robust feature representation for fake image detection. 

\begin{table}[h!]
\centering
\begin{tabular}{ |c|c|c|c| }
\hline
\textbf{Dataset} & \textbf{DFT} & \textbf{DWT} & \textbf{DFT + DWT} \\ \hline
StarGAN \cite{choi2018stargan} & 78.81\%  & 79.34\%  & 97.67\% \\ \hline
StyleGAN \cite{karras2019style} & 61.45\% & 69.11\% & 87.1\% \\ \hline
SITD \cite{chen2018learning} & 83.61\% & 49.72\% & 90.3\% \\ \hline
BigGAN \cite{brock2018large} & 64.92\% & 66.92\% & 72.08\% \\ \hline
StyleGAN2 \cite{karras2019analyzing} & 68.02\%  & 59.54\% & 91.83\% \\ \hline
CycleGAN \cite{zhu2017unpaired} & 65.31\% & 55.94\% & 77.53\% \\ \hline
Whichfaceisreal \cite{whichfaceisreal} & 39.85\% & 77.40\% & 80.15\% \\ \hline
GauGAN  \cite{Park_2019_CVPR} & 50.91\% & 67.22\% & 70.91\% \\ \hline
Deepfake  \cite{rossler2019faceforensics++} & 56.39\% & 51.88\% & 68.90\% \\ \hline
CRN \cite{chen2017photographic} & 61.75\% & 83.85\% & 55.06\% \\ \hline
IMLE \cite{li2019diverse} & 47.82\% & 79.90\% & 55.06\% \\ \hline
SAN \cite{dai2019second} & 69.18\% & 46.80\% & 49.32\% \\ \hline
\end{tabular}
\caption{ Evaluation of DFT and DWT combination for fake image detection. Percentage accuracy is reported for the full image using only DFT, only DWT, and the combination DFT+DWT.}
\label{table:dft_dwt_results}
\end{table}

\subsection{The Effect of Block Size}

We study the effect of using different block sizes instead of computing DFT over the whole image. In Table \ref{table:blocksize_results}, we have shown results of computing DFT on the block size of $8\times8$, $16\times16$, $32\times32$, and $256\times256$ (Full-Image size). Note that, block size experiments are performed by keeping exactly the same experimental settings. Results demonstrate that 8$\times$8 block size has consistently outperformed other block sizes. Therefore, transforming the image to the frequency domain using 8$\times$8 blocks for DFT  is more effective for fake image detection.
\subsection{The Effect of Color-Space}
We evaluate the effectiveness of converting images into  YCbCr color space before frequency transformations. We performed two experiments using the same settings to compare the performance of RGB with  YCbCr color space.
Results in Table \ref{table:colorspace_compar_results} show that converting an image to YCbCr colorspace adds more discriminative features in the frequency domain and helps in better fake image detection.

\begin{table}[h!]
\centering
\begin{tabular}{ |c|c|c|c|c| }
\hline
\textbf{Dataset} & \textbf{8x8} & \textbf{16x16} & \textbf{32x32} & \textbf{`Full-Image'} \\ \hline
StarGAN \cite{choi2018stargan} & $\textbf{94.62\%}$ & $77.74\%$ & $51.05\%$ & $78.81\%$ \\ \hline
StyleGAN \cite{karras2019style} & $\textbf{90.65\%}$ & $68.93\%$ & $46.47\%$ & $61.45\%$ \\ \hline
SITD \cite{chen2018learning} & $\textbf{86.1\%}$ & $55.28\%$ & $75.56\%$ & $83.61\%$ \\ \hline
BigGAN \cite{brock2018large} & $\textbf{69.42\%}$ & $63.12\%$ & $48.25\%$ & $64.92\%$ \\ \hline
StyleGAN2 \cite{karras2019analyzing} & $\textbf{92.23\%}$ & $65.26\%$ & $53.26\%$ & $68.02\%$ \\ \hline
CycleGAN \cite{zhu2017unpaired} & $\textbf{79.96\%}$ & $67.97\%$ & $36.68\%$ & $65.31\%$ \\ \hline
Whichfaceisreal \cite{whichfaceisreal} & $\textbf{81.50\%}$ & $63.25\%$ & $41.95\%$ & $39.85\%$ \\ \hline
GauGAN \cite{Park_2019_CVPR} & $\textbf{67.11\%}$ & $54.74\%$ & $57.95\%$ & $50.91\%$ \\ \hline
Deepfake \cite{rossler2019faceforensics++} & $\textbf{60.61\%}$ & $51.90\%$ & $57.35\%$ & $56.39\%$ \\ \hline
CRN \cite{chen2017photographic} & $51.43\%$ & $50.44\%$ & $57.90\%$ & $\textbf{61.75\%}$ \\ \hline
IMLE \cite{li2019diverse} & $51.50\%$ & $50.53\%$ & $\textbf{66.28}\%$ & $47.82\%$ \\ \hline
SAN \cite{dai2019second} & $46.58\%$ & $51.83\%$ & $\textbf{72.15\%}$ & $69.18\%$ \\ \hline

\end{tabular}
\caption{Fake image detection accuracy variation by varying block sizes for DFT transform. The block size 8$\times$8 has produced best results therefore in our experiments this block size is used.}
\label{table:blocksize_results}
\end{table}





\begin{table}[h!]
\centering
\begin{tabular}{ |c|c|c| }
\hline
\textbf{Dataset} & \textbf{RGB} & \textbf{YCbCr} \\ \hline
StarGAN \cite{choi2018stargan} & $66.83\%$  & $\textbf{78.81\%}$ \\ \hline
StyleGAN \cite{karras2019style} & $59.32\%$ & $\textbf{61.45\%}$ \\ \hline
SITD \cite{chen2018learning} & $\textbf{87.50\%}$ & $83.61\%$ \\ \hline
BigGAN \cite{brock2018large} & $\textbf{72.97\%}$ & $64.92\%$ \\ \hline
StyleGAN2 \cite{karras2019analyzing} & $60.42\%$  & $\textbf{68.02\%}$ \\ \hline
CycleGAN \cite{zhu2017unpaired} & $\textbf{75.89\%}$ & $65.31\%$ \\ \hline
Whichfaceisreal \cite{whichfaceisreal} & $47.80\%$ & $39.85\%$ \\ \hline
GauGAN  \cite{Park_2019_CVPR} & $\textbf{60.92\%}$ & $50.91\%$ \\ \hline
Deepfake  \cite{rossler2019faceforensics++} & $56.02\%$ & $\textbf{56.32\%}$ \\ \hline
CRN \cite{chen2017photographic} & $58.83\%$ & $\textbf{61.75\%}$ \\ \hline
IMLE \cite{li2019diverse} & $48.24\%$ & $47.82\%$ \\ \hline
SAN \cite{dai2019second} & $66.89\%$ & $\textbf{69.18\%}$ \\ \hline
\end{tabular}
\caption{The compassion of fake detection performance using RGB and YCbCr Colorspace. YCbCr color space has performed better than RGB color space.}
\label{table:colorspace_compar_results}
\end{table}

\section{Conclusions}
\label{sec:con}
This paper addresses the problem of fake image detection. For this purpose, a two-stream network is proposed consisting of a spatial stream and a frequency stream. The proposed method generalizes to unseen fake image generator distributions much better than the current state-of-the-art approaches. The proposed method is also found to be more robust to the common image perturbations including blur and JPEG compression artifacts. The improved performance is leveraged by combining two types of frequency domain transformations, namely, Discrete Fourier Transform (DFT) and Discrete Wavelet Transform (DWT). Both transformations are applied upon YCbCr color-space and different frequency domain channels are concatenated to discriminate fake images from the real ones. By exploiting the differences between the real and the fake image frequency responses, improved fake detection performance is achieved.  In the future, we aim to extend this work for fake video and audio detection.


\bibliographystyle{unsrt}
\bibliography{main}

\end{document}